\documentclass[11pt,a4paper]{article}
\usepackage[hyperref]{emnlp-ijcnlp-2019}
\usepackage{xcolor}
\usepackage{times}
\usepackage{latexsym}
\usepackage{tikz}
\usetikzlibrary{fit, positioning}
\usepackage{lipsum}
\usepackage{booktabs}
\usepackage{algorithm, algpseudocode}
\usepackage{multirow}
\usepackage{caption}
\usepackage{amsmath, amssymb, amsthm}
\usepackage{url}
\usepackage{pgfplots}
\usepackage{rst}
\usepackage{graphics}
\usepackage[flushleft]{threeparttable}

\aclfinalcopy 

\DeclareMathOperator*{\argmax}{argmax}
\DeclareMathOperator*{\argmaxk}{argmax_\emph{k}}

\expandafter\def\expandafter\normalsize\expandafter{%
    \normalsize
    \setlength\abovedisplayskip{5pt}
    \setlength\belowdisplayskip{5pt}
    \setlength\abovedisplayshortskip{5pt}
    \setlength\belowdisplayshortskip{5pt}
}
\algnewcommand\algorithmicswitch{\textbf{switch}}
\algnewcommand\algorithmiccase{\textbf{case}}
\algnewcommand\algorithmicassert{\texttt{assert}}
\algnewcommand\Assert[1]{\State \algorithmicassert(#1)}%
\algdef{SE}[SWITCH]{Switch}{EndSwitch}[1]{\algorithmicswitch\ #1\ \algorithmicdo}{\algorithmicend\ \algorithmicswitch}%
\algdef{SE}[CASE]{Case}{EndCase}[1]{\algorithmiccase\ #1}{\algorithmicend\ \algorithmiccase}%
\algtext*{EndSwitch}%
\algtext*{EndCase}%

\title{Neural Generative Rhetorical Structure Parsing}

\author{Amandla Mabona \\
    Dept. of Computer Science and Technology\\
    University of Cambridge\\
    \texttt{amandla.mabona@cl.cam.ac.uk}\\
  \And
  Laura Rimell \\
  DeepMind\\
  London, UK\\
  \texttt{laurarimell@google.com}\\
  \AND
  Stephen Clark\\
  DeepMind\\
  London, UK\\
  \texttt{clarkstephen@google.com}\\
  \And
  Andreas Vlachos \\
  Dept. of Computer Science and Technology\\
  University of Cambridge\\
  \texttt{andreas.vlachos@cst.cam.ac.uk}
}

\date{}

\begin{document}
\maketitle

\begin{abstract}
  Rhetorical structure trees 
  have been shown to be useful for several document-level tasks including summarization and document classification.
  Previous approaches to RST parsing have used discriminative
  models; however, these are less sample efficient than generative
  models, and RST parsing datasets are typically small.
  In this paper, we present the first generative model for RST
  parsing. Our model is a document-level RNN grammar (RNNG) with a bottom-up traversal order. We show that, for our parser's traversal order, previous
  beam search 
  algorithms for RNNGs have a left-branching 
  bias which is ill-suited for RST parsing.
  We develop a 
  novel beam search algorithm that keeps track of both structure- and word-generating actions
  without exhibiting this branching
  bias and results in absolute improvements of 6.8 and 2.9 on unlabelled and labelled  F1 over previous algorithms.
  Overall, our generative model outperforms a discriminative model with the same features
  by 2.6 F1 points and achieves  performance comparable to the state-of-the-art,
  outperforming all published parsers  from a recent replication study that do not use additional training data.
\end{abstract}

\section{Introduction}
\begin{figure}[t]
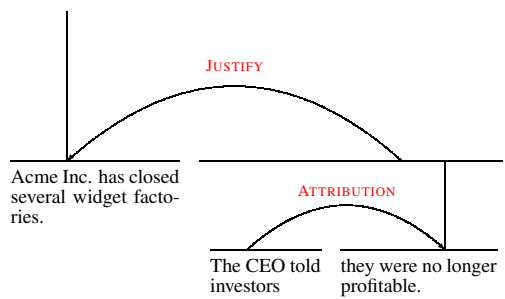

  \centering
  \scalebox{.68}{
  \dirrel{}
{\rstsegment{Acme Inc. has closed several widget factories.}}{Justify}
{\dirrel{Attribution}{\rstsegment{The CEO told investors}}{}{
    \rstsegment{they were no longer profitable.}}}}
  \caption{An example of an RST tree.}
  \label{fig:rst}
\vspace*{-0.3cm}
\end{figure}

Understanding a document's discourse-level organization is important for correctly interpreting it, and 
discourse analyses have been shown to be helpful for several NLP tasks \cite{bhatia_better_2015, ji_neural_2017, feng_patterns_2014, ferracane_leveraging_2017}. A popular formalism for discourse analysis is Rhetorical Structure Theory (RST) \cite{mann_rhetorical_1988} (Fig.~\ref{fig:rst}) which 
represents a document 
as a tree of discourse units recursively built by connecting smaller units
through rhetorical relations. Learning to predict RST trees is difficult because
it depends on pragmatics as well as literal meaning, and the English RST
Discourse Treebank (RST-DT) \cite{carlson_building_2003} is small by the
standards of modern parsing datasets, with 347 training documents.

Previous approaches to RST parsing \cite{ji_representation_2014, feng_linear-time_2014, joty_codra:_2015, braud_cross-lingual_2017} have used locally normalized discriminative
models. However, these are known
to have worse performance than generative models when there is little training
data \cite{ng_discriminative_2002, yogatama_generative_2017}. 

Unlike locally normalised discriminative models, generative models are not susceptible to label
bias \cite{lafferty2001conditional}. The success of generative
\cite{dyer_recurrent_2016, charniak2016parsing} and globally normalised
\cite{andor2016globally} syntactic parsers suggests that reducing label bias
leads to better performance. We hypothesize that using a generative parser would also lead to improved performance on RST parsing.
However, while they are free from label bias, generative parsers require more sophisticated search algorithms for decoding. \citet{fried_improving_2017} presented a word-level beam
search algorithm that made it possible  to decode directly from neural generative
parsers rather than using them as rerankers.

In this paper, we present the first generative RST parser\footnote{\citet{ji2016latent} introduced a neural generative discourse parser, but they used the annotation scheme of the Penn Discourse Treebank \cite{prasad2008penn} and Switchboard Dialog Act \cite{godfrey1992switchboard} corpora, predicting flat discourse representations between adjacent sentences, rather than hierarchical relations among clauses. }. Our model is a document-level version of an RNN Grammar (RNNG, 
\citet{dyer_recurrent_2016})
defined through a transition system with both word- and  structure-generating actions.  
 It uses distributed representations of
discourse units and transition probabilities parametrized by RNNs to model
unbounded dependencies in a document.

For our discourse parser, we find that \citet{fried_improving_2017}'s word-level beam search algorithm is biased towards producing left-branching trees.
We analyse the source of this bias and develop a novel 
beam search algorithm that removes it by tracking both word- and structure-generating actions.
On the  RST-DT development set, our  algorithm leads to improvements of 6.8\% and 2.9\% on
unlabelled and labelled attachment accuracies when decoding from the
same parser, compared to word-level beam search. On the  RST-DT test set, our generative parser outperforms a
  discriminative version with the same features by 2.6\% on labelled attachment accuracy.
  Overall, our parser obtains a labelled attachment score of 45.0\%,
  outperforming all published parsers in a recent replication study that do not use additional training data.

\section{Rhetorical Structure Theory}
\label{sec:rst-parsing}

Rhetorical Structure Theory describes the structure of a document in terms of text spans that form discourse units and the relations between them.
The basic unit of analysis is an \emph{elementary discourse unit} (EDU) which
can be assumed to be a syntactic clause. A \emph{unit} is made up
of two or more adjacent discourse units (which can be EDUs or other units) that are in some \emph{rhetorical relation}.

Most rhetorical relations are binary and asymmetric with one argument, the
\emph{nucleus}, being more important than the other, the \emph{satellite}.
\emph{Importance} is defined through a deletion test: a text becomes incoherent if
a nucleus is deleted, but not if a satellite is. These binary asymmetric relations are called
\emph{mononuclear} relations. The remaining relations are symmetric, having two or more arguments of equal importance, and are called \emph{multinuclear} relations.

An \emph{RST tree} or \emph{analysis} is a nested collection of discourse units that are either EDUs or units, where the top unit spans the whole text \cite{mann_rhetorical_1988}. RST parsing is the task of automatically predicting RST trees for documents.

\section{Rhetorical Structure RNNGs}
In this section, we present a generative model for predicting RST trees given a
document segmented into a sequence of EDUs $e_{1:m}$\footnote{As in most previous work on RST parsing, we use gold EDU segmentations in our experiments, but our parser would use the output of an EDU segmenter in practice.}.
The model is a
document-level RNNG in
bottom-up traversal order \cite{kuncoro_lstms_2018}. We first describe syntactic
RNNGs in section~\ref{sec:recurr-neur-netw}. We then describe our parser's transition system in section~\ref{sec:trans-sys}, and its transition model in section \ref{sec:trans-mod}.

\begin{table*}[t]
  \centering
  \begin{tabular}{cccccc}
    \toprule
    \textbf{Action} & \textbf{Before} & \textbf{After} &  \textbf{Probability} & \textbf{Condition}\\
    \midrule
     $\mathsf{GEN}(e)$ &$\langle S, B \rangle$&$\langle S|\mathsf{EDU}(e), B|e\rangle$& $p_{trans}(\mathsf{GEN}| S)\cdot p_{gen}(e|S)$ & $|B|<m$\\
     $\mathsf{RE}(r, n)$&$\langle S| U_L|U_R, B \rangle$&$\langle S|\big(\mathsf{Unit}(r, n)\ U_L\ U_R\big), B \rangle$&$p_{trans}(\mathsf{RE}(r, n)| S)$ & $|S|\geq 2$\\
    \bottomrule                       
  \end{tabular}
  \caption{Our transition system. $|S|$ is the number of discourse units on the
    stack, $|B|$ is the number of EDUs in the buffer and $m$ is the number of
    EDUs in the whole document, $r$ is a relation label and $n$ is a nuclearity label.}
  \label{tab:trans}
\end{table*}

\begin{table*}[t]
  \centering
  \begin{threeparttable}
  \begin{tabular}{ccc}
    \toprule
    \textbf{Stack} & \textbf{Buffer} & \textbf{Prediction}\\
    \midrule
    $\epsilon$ & $\epsilon$ & $\mathsf{GEN}(e_1)$\\
    $\mathsf{EDU}(e_1)$ & $e_1$ & $\mathsf{GEN}(e_2)$\\
    $\mathsf{EDU}(e_1)|\mathsf{EDU}(e_2)$ & $e_1|e_2$ & $\mathsf{GEN}(e_3)$\\
    $\mathsf{EDU}(e_1)|\mathsf{EDU}(e_2)|\mathsf{EDU}(e_3)$ & $e_1|e_2|e_3$ & $\mathsf{RE}(\mbox{\em ATTR}, \mbox{\em SN})$\\
    $\mathsf{EDU}(e_1)|(\mathsf{Unit}(\mbox{\em ATTR}, \mbox{\em SN})\ \mathsf{EDU}(e_2)\ \mathsf{EDU}(e_3))$ & $e_1|e_2|e_3$ & $\mathsf{RE}(\mbox{\em JUST}, \mbox{\em NS})$\\
    $\big(\mathsf{Unit}(\mbox{\em JUST}, \mbox{\em NS})\ \mathsf{EDU}(e_1)\ \big(\mathsf{Unit}(\mbox{\em ATTR}, \mbox{\em SN})\ \mathsf{EDU}(e_2)\ \mathsf{EDU}(e_3)\big)\big)$ & $e_1|e_2|e_3$ &\\

    \bottomrule
  \end{tabular}
   \begin{tablenotes}
   \item {\small $[e_1$ Acme Inc. has closed several widget factories. $]$ $[e_2$ The
     CEO told investors $]$ $[e_3$ they were no longer profitable. $]$}
   \end{tablenotes}
  \end{threeparttable}
  \caption{An example of a completed computation in our transition system.}
  \label{tab:computation}
  \vspace*{-0.3cm}
\end{table*}

\subsection{Recurrent Neural Network Grammars}
\label{sec:recurr-neur-netw}
Recurrent neural network grammars are a class of syntactic language models that
define a joint probability distribution $p(\boldsymbol{x}, \boldsymbol{y})$ over sentences and their phrase
structure trees. An RNNG is defined by a triple $(N, \Sigma, \Theta)$ with $N$ a finite set of nonterminal symbols, $\Sigma$ a finite set of terminal symbols and
$\Theta$ neural network parameters.

RNNGs generate sentences and their parse trees through actions\footnote{We use ``action'' and ``transition'' interchangeably.} in an
abstract state machine. A machine state is a tuple $\langle {S, B} \rangle$
where $S$ is a stack which holds
partial phrase structure trees and $B$ is a buffer which holds sentence prefixes.
The transitions push new subtrees onto the stack, combine subtrees already there,
and append terminals to the buffer until the stack contains a single phrase
structure tree and the buffer contains a complete sentence. The original
presentation in \citet{dyer_recurrent_2016} used the following transition system:
\begin{description}
\item[$\mathsf{NT}(X)$] Push the nonterminal node $(X$ onto the top of the stack, 
  where $X\in N$.
\item[$\mathsf{GEN}(w)$] Push the terminal symbol $w\in\Sigma$ onto the top of the stack
  and the end of the buffer. 
\item[$\mathsf{REDUCE}$] Pop subtrees $\tau_1, \cdots, \tau_l$ from the top of
  the stack until the first nonterminal node $(X$ is reached and push the
  subtree $(X \tau_1 \cdots \tau_l)$ onto the top of the stack. 
\end{description}

A sentence $\boldsymbol{x}$ and phrase structure tree $\boldsymbol{y}$ are
generated by a unique sequence of actions $a_{1:k}$. The joint distribution
$p(\boldsymbol{x}, \boldsymbol{y})$ is defined as the probability of the action
sequence $a_{1:k}$:

\begin{equation}
    p(\boldsymbol{x}, \boldsymbol{y}) = \prod_{j=1}^kp(a_{j}|a_{<j}) =
  \prod_{j=1}^kp(a_{j}|S_j, B_j)
\end{equation}

The next action distribution $p(a_j|S_j, B_j)$ is parametrized using neural
embeddings of the stack and buffer. Briefly, the next action distribution is computed using a softmax on the output
of a linear transformation on the state embedding, which is the concatenation of
a buffer embedding and a stack embedding. The
buffer embedding is the final hidden state of an LSTM that reads the word embeddings of
the words in the buffer. The stack embedding is the hidden state of a stack LSTM
that reads the embeddings of the subtrees on the stack. The embeddings of the
subtrees on the stack are computed recursively using a bidirectional LSTM that
reads the embeddings of the nonterminal symbol and its children.


\subsection{Transition System}
\label{sec:trans-sys}
We modify the RNNG
generative process so that it generates an EDU-segmented document and its RST tree.
In our model, the stack $S$ holds partial RST trees and the buffer $B$ holds a
sequence of EDUs (a prefix of a document's EDU segmentation). The
transitions generate EDUs and push them onto the stack and buffer, and combine RST subtrees
on the stack into new subtrees. The process terminates when the buffer contains
a complete document and the stack a single RST tree. 

\citet{kuncoro_lstms_2018} presented an RNNG variant with a bottom-up
transition system that replaces the $\mathsf{NT}(X)$ and $\mathsf{REDUCE}$
transitions with a single $\mathsf{REDUCE}(X, n)$ transition, as in traditional
shift-reduce parsers. In initial experiments, we found this variant outperformed
a model using the original top-down transition system. We hypothesize this
is because an RST non-terminal's label is more difficult to predict from its
parent's label than is the case in phrase structure trees, while a parent's label
can be predicted once its children have been seen.

Finally, RST trees are traditionally binarized so we modify the
$\mathsf{REDUCE}$ transitions accordingly, resulting in the following transition system (see also Table~\ref{tab:trans}): 

\begin{description}
\item[$\mathsf{GEN}(e)$] Generate the EDU $e$ and push it onto the top of the stack and the end of the buffer.
\item[$\mathsf{RE}(r, n)$] Pop the top two discourse units ($U_L$ and $U_R$) from
  the stack and push the unit $\big(\mathsf{Unit}(r, n)\ U_L\ U_R\big)$ onto the top
  of the stack, with $r$ and $n$  relation and nuclearity labels.
\end{description}

\noindent In our experiments, the relation labels $r$ are the 18 coarse-grained relations of \citet{carlson_discourse_2001}, while the nuclearity labels $n$ are in $\{\mbox{\em SN}, \mbox{\em NS}, \mbox{\em NN}\}$ corresponding respectively to a mononuclear relation with the nucleus on the right or the left and a binarized multinuclear relation.

Both transitions have conditions on when they can be performed (Table~\ref{tab:trans}). A \emph{computation} is a sequence of transitions where the condition for each transition is satisfied in its preceding state. A \emph{completed computation} for an input sequence is a computation where the final state buffer contains the input sequence and the final state stack contains a single tree. Table~\ref{tab:computation} shows an example of a completed computation for our transition system.

\subsection{Transition Model}
\label{sec:trans-mod}

In initial experiments we found, as did \citet{kuncoro_what_2017} for syntactic parsing, that conditioning only on the stack led to better
parsing accuracy, so we specify the next action distribution as
$p(a_j|S_j)$.
To handle the unbounded number of possible EDUs, we parametrize the
probabilities of $\mathsf{GEN}(e)$ actions using a neural language model. The next action distribution is factorised into a
structural action distribution $p_{trans}$ and a generation distribution
$p_{gen}$ as in \citet{buys_neural_2018}, so that $p(\mathsf{RE}(r, n)|S)=p_{trans}(\mathsf{RE}(r, n)|S)$ and 
$p(\mathsf{GEN}(e)|S)=p_{trans}(\mathsf{GEN}|S)\cdot p_{gen}(e|S)$ where
$p_{gen}$ is the neural language model. 

 We parametrize $p_{trans}$
 as a feedforward neural
 network on an embedding of the stack $\mathbf{h}_S(S)$. In initial experiments we found, consistent with \citet{morey_how_2017}, that a
 model with neural embeddings as its only features performed poorly. We
 therefore compute the representation using both neural embeddings of the discourse units on the stack (Section~\ref{sec:neural-stack}) and a set of structural features extracted from the stack (Section~\ref{sec:struc-feat}).

\subsubsection{Neural Embeddings}
\label{sec:neural-stack}
To produce the stack embedding, we first require embeddings for both EDUs and units.
We embed EDUs with bidirectional LSTMs\footnote{We track memory cells and use
  them when updating the hidden state in LSTMs and Tree LSTMs, but use only the
  hidden states for stack embeddings. Initial hidden states and memory cells are learned parameters.}. If $e$ is an EDU consisting of the word sequence $w_{1:k}$, then 
\begin{equation}
    \begin{split}
        &\mathbf{h}^{\rightarrow}_k = \mathsf{LSTM}^{(\rightarrow)}(\mathbf{w}_{1:k}, \mathbf{h}^{\rightarrow}_0)\\
        &\mathbf{h}^{\leftarrow}_k = \mathsf{LSTM}^{(\leftarrow)}(\mathbf{w}_{k:1}, \mathbf{h}^{\leftarrow}_0)
    \end{split}
\end{equation}
where $\mathbf{w}_t$ is the word embedding of $w_t$. The embedding for $e$, $\mathbf{h}_{EDU}(e)$, is
the concatenation of the final forward and backward hidden states:
\begin{equation}
  \mathbf{h}_{EDU}(e)=[\mathbf{h}^{\rightarrow}_k;\mathbf{h}^{\leftarrow}_k]  
\end{equation}

We embed units by composing their arguments with a Tree LSTM\footnote{Since constituency trees are $n$-ary branching, RNNGs for constituency parsing have used a
  bidirectional LSTM composition function
  \cite{dyer_recurrent_2016,kuncoro_what_2017,kuncoro_lstms_2018} to compose the
  variable number of children. RST trees are binarized so we do not need this
  feature.} \cite{teng_head-lexicalized_2017}. A Tree LSTM recursively composes
vectors while using memory cells to track long-term dependencies.  
We produce a new representation for each EDU $e$ by applying a linear
transformation to the EDU embeddings (omitting bias terms for brevity):
\begin{equation}
    \mathbf{h}_{U}(e)  = \mathbf{W}_{U,h}\cdot\mathbf{h}_{EDU}(e)
\end{equation}

   For a unit, we define the ``nuclear'' EDU of a
unit recursively as the nucleus if the nucleus is an EDU, or the nuclear EDU of the nucleus if the nucleus is itself a unit. 
For
multinuclear relations, we take the left-most nucleus. Then, if $\big(\mathsf{Unit}(r,
n)\ U_L\ U_R\big)$ is a unit and $e_N$ is its nuclear EDU, $\mathbf{h}_{EDU}(e_N)$ is the embedding of the nuclear EDU,
and $\mathbf{h}_R(r, n)$ is an embedding of the nuclearity-relation pair $(r, n)$ in  a lookup
table:
\begin{equation}
    \begin{split}
          \mathbf{h}_{U}(U) = \mathsf{TREELSTM}(&[\mathbf{h}_{EDU}(e_N); \mathbf{h}_R(r, n)],\\
            &\mathbf{h}_{U}(U_L), \mathbf{h}_{U}(U_R))
    \end{split}
\end{equation}
where $\mathbf{h}_{U}(U_L)$ and $\mathbf{h}_{U}(U_R)$ are the hidden state and memory cell of the left and right argument of the unit respectively.

We embed the stack with a stack LSTM \cite{dyer_transition-based_2015}. If the stack contents are $D_1|\cdots|D_m$ with each $D_i$ being a discourse unit, then
\begin{equation}
    \mathbf{h}^N_S(S) = \mathsf{LSTM}^S(\mathbf{h}_{U}(D_{1:m}), \mathbf{h}^S_0)
\end{equation}

\subsubsection{Structural Features}
\label{sec:struc-feat}
We 
extract additional features from the stack that have been found to be
useful in prior work.
%
As in \citet{braud_cross-lingual_2017}, for each discourse
unit, we extract the
word embeddings of up to three words whose syntactic head is not in the
unit, adding padding if there are fewer than three. We concatenate
these features for the top two discourse units on the stack, using a dummy
embedding if the stack only contains one discourse unit. We write
$\mathbf{h}^{head}_S(S)$ for these features.

We use a categorical feature for whether the top two discourse units are: 
in the same sentence; in different sentences; or  incomparable since one
of them spans multiple sentences. We also use an equivalent feature for
paragraphs. Feature values are represented by embeddings in 
a lookup table. We write $\mathbf{h}^{comp}_S(S)$ for these features.

Finally, we extract features describing the dominance relation
\cite{soricut_sentence_2003} between the top two discourse units on the stack.
If there is a word in one discourse unit whose syntactic head is in the other,
we extract the word embeddings of these two words as well as an embedding of the
dependency relation between them, otherwise we use a single dummy embedding. We write
$\mathbf{h}^{dom}_S(S)$ for these features.

The structural feature representation is then the concatenation of these three
features:
\begin{equation}
    \mathbf{h}^F_S(S) = [\mathbf{h}^{head}_S(S); \mathbf{h}^{comp}_S(S);
  \mathbf{h}^{dom}_S(S)]
\end{equation}

\noindent and the full stack representation is the concatenation of the neural embedding
and the feature representation:
\begin{equation}
    \mathbf{h}_S(S) = [\mathbf{h}^N_S(S); \mathbf{h}^F_S(S)]
\end{equation}

\subsubsection{Probability Distributions}
\label{sec:prob-distr}

The action distribution is parametrized using the stack representation and an MLP:
\begin{equation}
    \begin{split}
    p_{trans}(a|S) & = p_{trans}(a|\mathbf{h}_S(S))\\
                 & = \text{softmax}(\mathbf{W}_{trans}\cdot\mathbf{h}_S(S))
    \end{split}
\end{equation}

We parametrize the EDU generation distribution $p_{gen}(e|S)$ with an LSTM decoder:
\begin{equation}
\mathbf{h}^{DEC}_t = \mathsf{LSTM}^{DEC}(\mathbf{w}_t, \mathbf{h}^{DEC}_{t-1})
\end{equation}

If $e=w_{1:k}$ then

\algtext*{EndFor}
\algtext*{EndWhile}
\algtext*{EndFunction}
\algtext*{EndIf}
\algrenewcommand\algorithmicdo{}

\begin{figure*}[h!]
  \begin{minipage}{\columnwidth}
\begin{algorithm}[H]
    \caption{Word-level Beam Search}\label{alg:ws}
  \footnotesize
  \begin{algorithmic}[1]
\Function{Search}{$x_{1:m}, k$}
\State $\mathcal{B}[0, 0], \gets\{\big(1, (\epsilon, \epsilon)\big)\}$
\For{$i\gets$ \Call{range}{$0$, $m$}}
\State $j \gets 0$\label{algws:init}
\While{$|\mathcal{B}[i, j]|\geq 0$ \textbf{and} $|\mathcal{B}[i+1, 0]| < k$}\label{algws:while}
\For{$(v, s)\gets$ \Call{top}{$\mathcal{B}[i, j], b$}} 
\For{$(a, s')\gets$ \Call{succ}{$s$}}
\State $v'\gets v\cdot p(a|s)$
\If{\Call{Complete}{$s'$}}
  \State \Call{push}{$\mathcal{B}[m+1, 0]$, $(v', s')$}
\Else
\Switch{$a$}
\Case{$\mathsf{GEN}(e_{i+1})$}
\State \Call{push}{$\mathcal{B}[i+1, 0]$, $(v', s')$}
\EndCase
\Case{$\mathsf{RE}(r, n)$}
\State \Call{push}{$\mathcal{B}[i, j+1]$, $(v', s')$}
\EndCase
\EndSwitch
\EndIf
\EndFor
\EndFor
\State $j\gets j+1$\label{algws:inc}
\EndWhile
\EndFor
\State \textbf{return} \Call{top}{$\mathcal{B}[m+1, 0], 1$}
\EndFunction
\end{algorithmic}
\end{algorithm}  
\end{minipage}\hfill
\begin{minipage}{\columnwidth}
  \centering
  \resizebox{\columnwidth}{!}{
  \hspace*{-0.5cm}\begin{tikzpicture}
    \begin{axis}[
      ymin=0,
      ymax=1,
      xlabel=$n$,
      ylabel=$P_L(T)$
      ]
      \addplot plot coordinates {
        (10, 0.33)
        (20, 0.63)
        (40, 0.82)
        (80, 0.91)
        (160, 0.95)
        (320, 0.97)
      };
    \end{axis}
  \end{tikzpicture}}
  \caption{Median degree of left branching for trees obtained from a bottom-up RNNG with a uniform scoring
    model using word-level beam search for sequences of various lengths ($n$).}
  \label{fig:bias}
  \end{minipage}
\end{figure*}
\vspace{-0.3cm}

\begin{align}
  p_{gen}(e|S) &= p_{gen}(w_{1:k}|S) \\
  &= \prod_{t=1}^k p_{gen}(w_t|w_{<t},S)\\
                     &= \prod_{t=1}^k p_{gen}(w_t|\mathbf{h}^{DEC}_{t-1},\mathbf{h}_S(S))
\end{align}
where
\begin{align}
  &p_{gen}(w_t|\mathbf{h}^{DEC}_{t-1},\mathbf{h}_S(S))\\
  &= \text{softmax}(\mathbf{W}_{gen}\cdot[\mathbf{h}_S(S); \mathbf{h}^{DEC}_{t-1}])
\end{align}
\pgfplotsset{grid style={dashed}}

\section{Inference}
\label{sec:inference}

Our generative model specifies a joint probability $p(\boldsymbol{x},
\boldsymbol{y})$. We parse a document $\boldsymbol{x}$ by  finding the MAP tree 
$\boldsymbol{y}^*$:
\begin{align}
  \label{eq:map}
  \boldsymbol{y}^* &= \argmax_{\boldsymbol{y}\in\mathcal{Y}(\boldsymbol{x})}p(\boldsymbol{y}|\boldsymbol{x})\\
  &= \argmax_{\boldsymbol{y}\in\mathcal{Y}(\boldsymbol{x})}p(\boldsymbol{x}, \boldsymbol{y})
\end{align}

\noindent The search space grows exponentially with the input length, so we must perform
inexact search as our model conditions on the entire relation structure of every subtree on the stack. 

Search is generally more difficult for generative models than for
discriminative ones, requiring more complex search algorithms.
For this reason, \citet{dyer_recurrent_2016} used RNNGs only to rerank the output of a discriminative
parser. \citet{fried_improving_2017} presented the
first algorithm for decoding directly from RNNGs to give competitive performance. They found that action-level beam search \cite{zhang_tale_2008} gave
poor performance for constituency parsing with RNNGs. 
The problem was that $\mathsf{GEN}$ actions
  almost always have lower probabilities than structure-generating actions,
  causing computations where $\mathsf{GEN}$ actions come earlier to ``fall
  off the beam'' even if the completed computation would have a higher probability than
  other completed computations.

To address this problem, \citet{fried_improving_2017} proposed \emph{word-level beam search}
(Algorithm~\ref{alg:ws}). Briefly, the algorithm keeps an array of beams indexed by the
current position in the sequence and the number of structure-generating actions
taken since this position was reached. The first beam for the current position $\mathcal{B}(i, 0)$ is filled
from the successors of beams for the previous position $\mathcal{B}(i-1, j)$ (lines~\ref{algws:init} to~\ref{algws:inc})
starting with $\mathcal{B}(i-1, 0)$ (line~\ref{algws:init}) and incrementing $j$ (line~\ref{algws:inc}) until there are at least $k$
items in $\mathcal{B}(i, 0)$ (line~\ref{algws:while}). The intuition is that analyses with the smallest number of structural actions since the previous beam was pruned have priority on the current beam.

We applied \citet{fried_improving_2017}'s algorithm\footnote{We used \emph{candidate fast-tracking} as described in \citet{stern_effective_2017}'s extension to \citet{fried_improving_2017}'s algorithm.} to our model, but found it
was biased towards producing left-branching trees. This led to poor performance as the \emph{right-frontier
    constraint} \cite{polanyi1988formal,webber9discourse,asher2012reference,asher2003logics} suggests discourse trees should be generally right-branching. In the next section, we present an analysis of the source of this bias and a novel 
beam search algorithm that corrects it.

\subsection{Diagnosing Branching Bias}

The trees returned by a trained parser depend on both the (learned) scoring model and the search
algorithm.
We can isolate bias in search algorithms by studying the
trees they return when the scoring model contains no information.
Intuitively, if the scoring model has no preference over trees, then any
preference shown by the parser is the result of biases in the search algorithm.

We tested whether the left-branching bias came from the word-level beam search (the search algorithm of \citet{fried_improving_2017}) by
using it to parse sequences of various lengths using a bottom-up RNNG with a uniform scoring model. We
broke ties at beam cut-offs by uniform sampling without replacement.
We measured branching bias using \citet{sampson_depth_1997}'s
production-based measure of left-branching for parse trees which we write as
$P_L(T)$ for a tree $T$. The measure
is the fraction of non-terminals whose left child is also a non-terminal, and varies from 0 for a fully
right-branching tree to $\frac{n-2}{n-1}\to 1$ for a fully left-branching tree, where $n$ is the number of leaves.
Figure ~\ref{fig:bias} shows the median value of this measure for 100 trees
each for sequences of various lengths from our uniform scoring model. It shows substantial left-branching
bias which increases with sequence length.

\begin{figure*}[t!]
  \centering
  \begin{tabular}{l@{\hskip 0.75in}l}
    \begin{tikzpicture}
    [
    dot/.style={circle,draw=black, fill,inner sep=1pt},
    reddot/.style={circle,draw=black, very thick, fill=red!80,inner sep=1.5pt},
    graydot/.style={circle,draw=black,  very thick, fill=gray!80,inner sep=1.5pt},
    bluedot/.style={circle,draw=black, very thick, fill=blue!80,inner sep=1.5pt},
    ]
    \foreach \x in {0, ..., 6}{
    \foreach \y in {0, ..., 5}{
      \node[dot] at (\x, \y){ };
    }
  }

  \foreach \y in {1, ..., 4}{
    \node[graydot] at (6, \y){};
  }

    \foreach \x in {0, ..., 2}{
    \foreach \y in {0, ..., \x}{
      \draw[->, very thick, draw=gray, dotted] (\x + 3, \y + 1 + 0.1 ) -- (\x + 3, \y + 1 +  1 - 0.1);
      \draw[->, very thick, draw=gray, dotted] (\x + 3 + 0.1, \y + 1 + 1 - 0.1 ) -- (\x + 3  + 1- 0.1, 0.1) {};
    }
  }

  \foreach \x in {2, ..., 5}{
      \draw[->, very thick, draw=blue, loosely dashdotted] (\x, 0.1) -- (\x, 0.9);
      \draw[->, very thick, draw=blue, loosely dashdotted] (\x + 0.1, 0.9) -- (\x + 0.9, 0.1);
  }

  \foreach \x in {3, ..., 6}{
    \draw[->, very thick, draw=red, dashed] (\x - 0.9, 0) -- (\x - 0.1, 0);
  }

  \foreach \x in {1, ..., 5}{
    \draw[->, very thick, draw=red, dashed] (6, \x - 0.9) -- (6, \x - 0.1);
  }


\foreach \y in {1,...,5}{
  \node[below,xshift=-3mm, yshift=+3mm] at  (-.1,\y) {\y};
}

\foreach \x in {1,...,6}{
  \node[below,yshift=-1mm] at  (\x, -.1) {\x};
  }
  
\node[below,xshift=-2mm,yshift=-1mm] at (0,0) {0};

\node[bluedot] at (6,1){};      
\node[reddot] at (6,5){};      
\draw[->, very thick, draw=violet] (0.1,0) -- (0.9,0);
\draw[->, very thick, draw=violet] (6,0.1) -- (6,0.9);
\draw[->, very thick, draw=violet] (1.1,0) -- (1.9,0);
\end{tikzpicture}
&
  \begin{tikzpicture}
    [
    dot/.style={circle,draw=black, fill,inner sep=1pt},
    reddot/.style={circle,draw=black, very thick, fill=red!80,inner sep=1.5pt},
    bluedot/.style={circle,draw=black, very thick, fill=blue!80,inner sep=1.5pt},
    violetdot/.style={circle,draw=black, very thick, fill=violet!80,inner sep=1.5pt}
    ]

    \foreach \x in {0, ..., 6}{
    \foreach \y in {0, ..., 5}{
      \node[dot] at (\x, \y){ };
    }
  }

  \foreach \x in {2, ..., 5}{
    \draw[->, very thick, draw=blue, loosely dashdotted] (\x, \x - 1.9) -- (\x, \x - 1.1); 
    \draw[->, very thick, draw=blue, loosely dashdotted] (\x + 0.1, \x - 1) -- (\x + 0.9, \x - 1); 
  }
  \foreach \z in {1, ..., 3}{
    \foreach \x in {\z, ..., 3}{
    \draw[->, very thick, draw=gray, dotted] (\x + 3 - 0.9, \z) -- (\x +3 - 0.1, \z); 
  }
  \foreach \x in {1, ..., \z}{
    \draw[->, very thick, draw=gray, dotted] (\z - 1 + 3, \x - 1 + 0.1) -- (\z -1 + 3 , \x -1 + 0.9); 
  }
}

    \draw[->, very thick, draw=gray, dotted] (3, 0.1) -- (3 , 0.9); 
  \foreach \x in {2, ..., 5}{
    \draw[->, very thick, draw=red, dashed] (\x + .1,0) -- (\x + 0.9,0); 
  }
  \foreach \x in {0, ..., 3}{
    \draw[->, very thick, draw=red, dashed] (6, \x + .1) -- (6, \x + 0.9); 
  }

\node[below,xshift=-2mm,yshift=-1mm] at (0,0) {0};

\foreach \y in {1,...,5}{
  \node[below,xshift=-3mm, yshift=+3mm] at  (-.1,\y) {\y};
}

\foreach \x in {1,...,6}{
  \node[below,yshift=-1mm] at  (\x, -.1) {\x};
  }

\node[dot] at (5,1){};  
\node[dot] at (5,0){};
\node[violetdot] at (6, 5){};
\draw[->, very thick, draw=violet] (0.1,0) -- (0.9,0);
\draw[->, very thick, draw=violet] (1.1,0) -- (1.9,0);
\draw[->, very thick, draw=violet] (6,4.1) -- (6,4.9);
\end{tikzpicture}\\

\end{tabular}
 \caption{Word-level and bag-level beam search (left and right respectively) for
   an input sequence with 6 tokens. Nodes represent beams and paths represent
   computations. The horizontal axis is the number of $\mathsf{GEN}$
   actions and the vertical axis is the number of $\mathsf{RE}$ actions for
   bag-level search and the number of $\mathsf{RE}$ actions since the last
   $\mathsf{GEN}$ for word-level search. We show the path of a left-branching tree in blue with dashed and dotted lines and a
   right-branching tree in red with dashed lines. We show possible transitions between beams that do not belong to either of these paths in gray with dotted lines. Red, blue and purple dots respectively show the beam where
   the computation of a right-branching tree, left-branching tree or both
   are completed.}                   
  \label{fig:path}
\end{figure*}
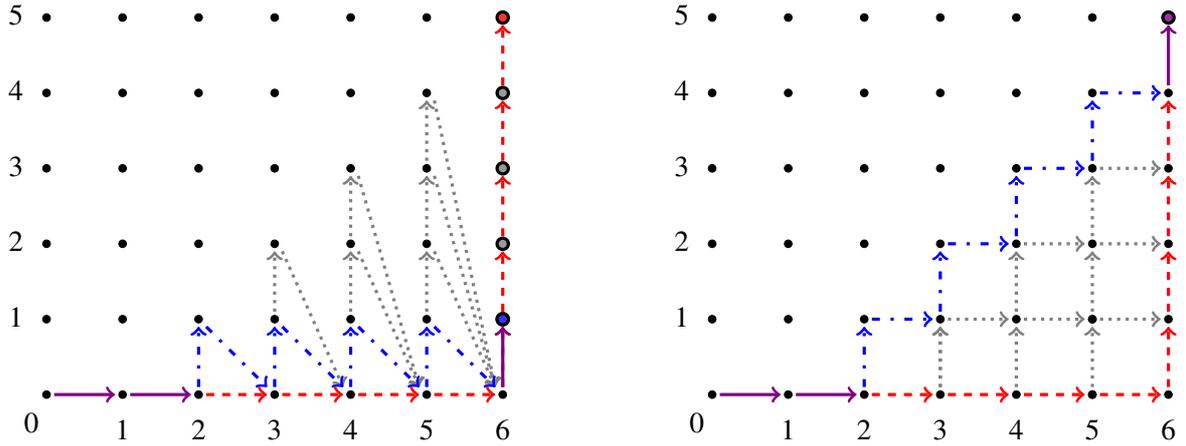
Word-level beam search has two sources of bias: first, computations with
fewer $\mathsf{RE}$ actions since the last $\mathsf{GEN}$ action are added to the next beam
first (lines~\ref{algws:init} and~\ref{algws:inc} in Alg.~\ref{alg:ws}). Computations with more actions are only considered if the next
beam is not already full by the time they are reached (line~\ref{algws:while} in Alg.~\ref{alg:ws}). This means the next beam may fill
up before these computations are even considered and they will ``fall off
the beam''. A right-branching subtree over $k$ leaves has $k$ consecutive
$\mathsf{GEN}$ actions followed by $k-1$
consecutive $\mathsf{RE}$ actions, meaning it results in a computation in $\mathcal{B}(i_k, k-1)$
where $i_k$ is the position of the $k$-th leaf. Thus right-branching subtrees are
later in line to be considered and are increasingly likely to fall off the beam as
they span more leaves.

Second, the beams $\mathcal{B}(i, j)$ contain computations with unequal
numbers of actions. For a binary tree with $m$ leaves, all completed computations have $m$ $\mathsf{GEN}$ and $m-1$ $\mathsf{RE}$ actions. The total number of actions up to the $k$-th $\mathsf{GEN}$ action varies, though, from $k$ to $2k-2$. Since the probability of a computation $a_{1:l}$ is $\prod_{j=1}^l p(a_j|a_{<j})$, this means word-level beam search compares computations 
with different numbers of factors contributing to their probabilities. This bias does not necessarily favour left-branching trees, but it does introduce a potential problem when comparing computations.

\subsection{Bag-Level Beam Search}
\label{sec:dyn-prog}

\algtext*{EndFor}
\algrenewcommand\algorithmicdo{}

We now present a beam search algorithm without these sources of bias (Algorithm~\ref{alg:bsbeam}). Our
algorithm is based on a simple dynamic program that keeps track of the number of
$\mathsf{GEN}$ and $\mathsf{RE}$ actions separately. This (i) allows us to consider computations from all source beams simultaneously and (ii) ensures all computations in a beam have the same number of actions. Since this is equivalent to keeping separate beams for different bags of
unlabelled actions, we call the algorithm bag-level beam search.

\begin{algorithm}[t]
\caption{Bag-level Beam Search}\label{alg:bsbeam}
\footnotesize\begin{algorithmic}[1]
\Function{Search}{$x_{1:m}, k$}
\State $B[0, 0]\gets\{\big(1, (\epsilon, \epsilon)\big)\}$
\For{$i\gets$ \Call{range}{$0$, $m$}}\Comment{$\mathsf{GEN}(e_i)$}
\For{$j\gets$ \Call{range}{$0$,  $i-1$}}\Comment{$\mathsf{RE}(r, n)$}\label{algws:loop}
\For{$(v, s)\gets$ \Call{top}{$B[i, j], k$}} 
\For{$(a, s')\gets$ \Call{succ}{$s$}}
\State $v'\gets v\cdot p(a|s)$
\Switch{$a$}
\Case{$\mathsf{GEN}(e_{i+1})$}
\State \Call{push}{$B[i+1, j]$, $(v', s')$}
\EndCase
\Case{$\mathsf{RE}(r, n)$}
\State \Call{push}{$B[i, j+1]$, $(v', s')$}
\EndCase
\EndSwitch
\EndFor
\EndFor
\EndFor
\EndFor
\State
\State \textbf{return} \Call{top}{$B[m, m-1], 1$}
\EndFunction
\end{algorithmic}
\end{algorithm}

We write $\mathcal{C}(i, j)$ for the set of computations with $i$ $\mathsf{GEN}$ actions and $j$ $\mathsf{RE}$ actions. Then all completed computations are in $\mathcal{C}(m, m-1)$ for an input sequence of length $m$. 

For each computation $c\in\mathcal{C}(i, j)$, the last action was either a $\mathsf{GEN}$ or an $\mathsf{RE}$ action, so $c$ 
is either of the form $c=\mathsf{GEN}|c^{'}$ where 
$c'\in\mathcal{C}(i-1, j)$ or it is of the form $c=\mathsf{RE}|c^{''}$ where $c^{''}\in\mathcal{C}(i, j-1)$.

The highest scoring computation in $\mathcal{C}(i, j)$, $c^*(i, j)=    \displaystyle\argmax_{c\in\mathcal{C}(i, j)} p(c)$, is then the highest scoring computation ending on a $\mathsf{GEN}$ or an $\mathsf{RE}$\footnote{We omit actions' parameters for conciseness.}:

\begin{small}
\begin{align}
    c^*(i, j) = 
    \argmax \left\{p(c)\Bigg| 
    \arraycolsep=0.4pt\begin{array}{ll}
     c=\mathsf{GEN}|c^{'}, c^{'}\in\mathcal{C}(i-1, j);   \\
     c=\mathsf{RE}|c^{''}, c^{''}\in\mathcal{C}(i, j-1)
    \end{array}
    \right\}
\end{align}
\end{small}

There are exponentially many computations in $\mathcal{C}(i, j)$ so taking exact
maxima is intractable. Therefore we only take maxima over beams
$\mathcal{B}(i, j)$ which we update according to 

\begin{small}
\begin{align}
    \mathcal{B}(i, j) = 
    \argmaxk \left\{p(c)\Bigg| 
    \arraycolsep=0.4pt\begin{array}{ll}
     c=\mathsf{GEN}|c^{'}, c^{'}\in\mathcal{B}(i-1, j);   \\
     c=\mathsf{RE}|c^{''}, c^{''}\in\mathcal{B}(i, j-1)
    \end{array}
    \right\}
\end{align}
\end{small}

\noindent where, in the set notation, ``;'' means ``or''.

We perform this recursive calculation for all $i$ and $j$ using the
dynamic program in Algorithm~\ref{alg:bsbeam}. 

Figure~\ref{fig:path} shows the differences between word-level and
bag-level beam search with example trajectories through the array of beams for
computations corresponding to a
left-branching (blue, dashed and dotted) and a right-branching (red, dashed) tree. Each path through the
lattice from $(0, 0)$ to $(i, j)$ defines a computation and shows the beams
it must pass through to end up in $\mathcal{B}(i, j)$.
 
The path length is equal to the number of actions taken to reach $(i, j)$.
In word-level beam search, paths
through beams with more consecutive $\mathsf{RE}$ actions (higher values on the vertical axis)
are only explored if the next word beam is
not already full (lines~\ref{algws:init},~\ref{algws:while} and \ref{algws:inc} in Algorithm~\ref{alg:ws}). This means the final beam may be full before the red path is
considered, causing it to ``fall off the beam''. In bag-level beam search, paths into a beam from both source beams
are considered and pruned simultaneously.
This addresses the first source of bias, namely that sequences with fewer consecutive $\mathsf{RE}$ actions are given priority.
All paths to
$(i, j)$ also have the same length; in other words, all computations in $\mathcal{B}(i, j)$ have the same
number of  actions ($i+j$), addressing the second source of bias.
\begin{table}[]
    \centering
    \resizebox{\columnwidth}{!}{%
  \begin{tabular}{cccccc}
    \toprule
    \multirow{2}{*}{\textbf{Metric}} &  \multirow{2}{*}{\textbf{Algorithm}} & \multicolumn{4}{c}{\textbf{Beam Size}} \\
                                     & & 10 & 20 & 40 & 80 \\
    \midrule
    \multirow{2}{*}{S} & Word-level Search & 58.3 & 58.4 & 58.4 & 60.8\\
                                     &  Bag-level Search & 66.4 & 67.3 & 67.0 & 67.6\\
    \midrule
    \multirow{2}{*}{N} &  Word-level Search & 50.3 & 50.5 & 50.5 & 51.8\\
                                     &  Bag-level Search & 56.1 & 56.4 & 56.2 & 56.9 \\
    \midrule
    \multirow{2}{*}{R} &  Word-level Search & 43.1 & 43.2 & 43.2 & 43.7\\
                                     &  Bag-level Search & 45.4 & 46.8 & 46.5 & 46.9\\
    \midrule
    \multirow{2}{*}{F} &  Word-level Search & 42.0 & 42.3 & 42.3 & 42.9\\
                       &  Bag-level Search & 44.7 & 45.5 & 45.3 & 45.8 \\
    \bottomrule
  \end{tabular}}
\caption{Dev. set micro-averaged $F_1$ scores on labelled attachment
  for word-level and bag-level beam search.}
  \label{tab:beam}
\end{table}
\section{Experiments}
\begin{table*}[t!]
   \centering
  \begin{tabular}{ccccc}
    \toprule
    \textbf{Model} &  \textbf{S}   &\textbf{N} &\textbf{R} &\textbf{F}\\
    \midrule
    \multicolumn{5}{l}{\emph{Feature-based parsers}}\\
     \citet{hayashi_empirical_2016} & 65.1 & 54.6 & 44.7 & 44.1\\
    \citet{surdeanu_two_2015} & 65.3 & 54.2 & 45.1 & 44.2\\
    \citet{joty_codra:_2015} & 65.1 & 55.5 & 45.1 & 44.3\\
    \citet{feng_linear-time_2014} & \emph{\textbf{68.6}} & \emph{55.9} & \emph{\textbf{45.8}} & \emph{44.6}\\
    \midrule
    \multicolumn{5}{l}{\emph{Neural parsers}}\\
    \citet{braud_multi-view_2016} & 59.5 & 47.2 & 34.7 & 34.3\\
    \citet{li_discourse_2016} & 64.5 & 54.0 & 38.1 & 36.6\\
    \citet{braud_cross-lingual_2017} (mono) & 61.9 & 53.4 & 44.5 & 44.0\\
    \multicolumn{5}{l}{\emph{Our work}}\\
    Discriminative Baseline & 65.2 & 54.9 & 42.8 & 42.4\\
    Generative Model & \emph{67.1} & \emph{\textbf{57.4}} & \emph{45.5} & \emph{\textbf{45.0}} \\
    \midrule
    \multicolumn{5}{l}{\emph{Unpublished}}\\
    \citet{ji_representation_2014} (updated) & 64.1 & 54.2 & 46.8 & 46.3\\
    \multicolumn{5}{l}{\emph{Additional data}}\\
    \citet{braud_cross-lingual_2017} (cross + dev) & 62.7 & 54.5 & 45.5 & 45.1\\
    \bottomrule
  \end{tabular}
    
  \caption{Test set micro-averaged $F_1$ scores on labelled attachment decisions. We report numbers for other parsers from
    \citet{morey_how_2017}'s replication study. For each metric, the highest score for all the parsers in the comparison is shown in bold, while the highest score among parsers of that type (neural or feature-based) is in italics.}
  \label{tab:gold}
 \vspace{-0.2cm}
\end{table*}
\subsection{Dataset}
\label{sec:data}

We train and evaluate our models on the RST Discourse Treebank (RST-DT)
\cite{carlson_discourse_2001}. We evaluate on the standard test set,
but we use 25 documents from the training set as a development set. To reduce the
rare token count, we use the spaCy \cite{spacy2} named entity recognition model to replace
named entities with their named entity tags. 

\subsection{Evaluation}
We follow the evaluation setup used by \citet{morey_how_2017}. They performed a
replication study of several competitive RST parsers and implemented a consistent evaluation procedure. They found that micro-
and macro-averaged $F_1$ had been used inconsistently in the RST parsing
literature, and that the standard evaluation metrics (RST Parseval) gave inflated
results. Following this study we evaluate using micro-averaged $F_1$ scores on
labelled attachment decisions as calculated by the EDUCE python package\footnote{https://github.com/irit-melodi/educe}. We report $F_1$ for predicting span attachments (S), span attachments with nuclearity (N), span attachments with relation labels (R) and span attachments with nuclearity and relation labels (F). 

We compare our results against the numbers from \citet{morey_how_2017}, since they include several competitive parers under a consistent evaluation scheme.\footnote{We do not compare against \citet{yu2018transition}, \citet{zhang2018rst} and \citet{lin2019unified}'s recent neural RST parsers since they do not evaluate labelled attachment decisions so their results are not comparable to ours.}

As a baseline, we use a discriminative version of our model. This is a
shift-reduce parser with the same EDU, unit and stack representations as our
model, but with a lookahead buffer representation as well. For the buffer
representation, we run a backward LSTM over the representations of the
remaining EDUs in the buffer.

\subsection{Training and Hyperparameters}
\label{sec:train-hyper}
We use 300-dimensional word embeddings initialized to word2vec
vectors \cite{mikolov_distributed_2013}. We tie the embeddings in the EDU LSTM and the decoder LSTM input and output
embeddings. We use a 2-layer bidirectional LSTM with 512-dimensional hidden
state for the EDU LSTM. The TreeLSTM composition function also has a 512-dimensional (in total) hidden
state with 100-dimensional relation embeddings. The stack LSTM and decoder LSTM
also have 512-dimensional hidden states. For the structural features, we use 10-dimensional sentence and paragraph boundary feature embeddings and 50-dimensional dependency relation embeddings.

We train the models with Adam \cite{kingma_adam:_2014} using an initial learning rate of
$10^{-3}$ and default values for the other hyperparameters. 
%
We apply blank noise variational smoothing \cite{kong_variational_2019} with a dropout rate of 0.25 to the tied embeddings
to regularize the model. In particular, for each document we sample a set of word
types to drop and replace their word embeddings with the $\mathtt{<UNK>}$ token's word embedding.  

We extract structural features using the sentence and paragraph boundary
annotation in the RST-DT, and dependency trees obtained from the spaCy parser.
Our models were implemented
in PyTorch \cite{paszke_automatic_2017}.

\subsection{Results}
\subsubsection{Search Comparison}
\label{sec:rst-parsing}

Table~\ref{tab:beam} shows RST-DT development set labelled attachment metrics
for our parser using word-level and bag-level beam search. Our search algorithm
outperforms word-level beam search on all of the metrics across beam
sizes.\footnote{Word-level beam search has three beam size parameters: the structural action beam size $k$, the next-word
  beam size $k_w$  and the fast-track beam
  size $k_s$. For top-down parsers, \citet{stern_effective_2017} set $k_w=k/10$, $k_s=k/100$, but in
  tuning these parameters on the development set we found the best performance with
  $k_w=k$, $k_s=k/10$ for  our parser.} 
On spans with nuclearity (N), bag-level beam search outperforms word-level beam search by 5.9\% to 8.1\%.
This is consistent with the branching bias in word-level search leading it to
return trees whose structure differs from the trees in the RST-DT. The poor
performance on structure prediction also seems to have a knock-on effect on the relation
and full tree prediction accuracy. 

\subsubsection{Parsing Performance}
\label{sec:parsing-results}
Table~\ref{tab:gold} shows RST-DT test set labelled attachment metrics for
various parsers. Our model outperforms all of the published\footnote{\citet{ji_representation_2014} presented a transition-based parser that used continuous bag-of-words representations for EDUs and an SVM as the next action classifier. For \citet{morey_how_2017}'s study, they submitted predicted discourse trees from an updated, unpublished version of their parser.} neural models that
do not use  additional training data\footnote{In the cross+dev setting, \citet{braud_cross-lingual_2017} train their parser on RST discourse treebanks for several languages.} in \citet{morey_how_2017}'s replication
study on all of the metrics. On span accuracy (S), we outperform all of the other
parsers except for \citet{feng_linear-time_2014}'s graph CRF model. On spans with
nuclearity (N), the equivalent of the unlabelled attachment score for discourse
dependencies, we outperform all of the parsers in the study. We perform
competitively on spans with relations (R), and we outperform all of the
published parsers that do not use additional data on spans with nuclearity
and relations (F).

 Our model also outperforms the discriminative baseline using the same features
 and implementation on all metrics by between 1.9\% and 2.7\%.

\section{Conclusion}

We introduced the first generative model for RST
parsing. We showed that word-level beam search has a branching bias for bottom-up RNNGs which hurt
performance on our task. We proposed a novel beam search algorithm that does not
have this branching bias and that outperformed word-level beam search across beam sizes and with different evaluation metrics. With our search algorithm, our generative model achieved
state-of-the art-level RST parsing performance, outperforming all of the
published RST parsers from a recent study that do not use additional training data on labelled attachment $F_1$. Our results show that generative modelling is an effective approach to RST
parsing, with superior structure prediction and competitive relation prediction performance.

\section*{Acknowledgments}
Andreas Vlachos is supported by the EPSRC grant eNeMILP
(EP/R021643/1). Amandla Mabona is supported by Commonwealth and Sansom Scholarships. We thank the anonymous reviewers for their helpful comments.

\bibliography{emnlp_rst}
\bibliographystyle{acl_natbib}
\appendix

\end{document}